\documentclass{article}

\usepackage[final]{neurips_2025}

\usepackage[utf8]{inputenc} 
\usepackage[T1]{fontenc}    
\usepackage{hyperref}       
\usepackage{url}            
\usepackage{booktabs}       
\usepackage{amsfonts}       
\usepackage{nicefrac}       
\usepackage{microtype}      
\usepackage{xcolor}         
\usepackage{amsmath}
\usepackage{bbm}
\usepackage{graphicx}
\usepackage{algorithm}
\usepackage{algpseudocode}
\usepackage{multirow}
\usepackage{pifont}
\usepackage{graphicx}
\usepackage{tabularx} 
\usepackage{booktabs}
\usepackage{pdfpages}

\usepackage{wrapfig}

\newcommand{\cmark}{\textcolor{black}{\ding{51}}} 
\newcommand{\xmark}{\textcolor{lightgray}{\ding{55}}}  

\definecolor{myorange}{HTML}{FFB570}
\definecolor{myblue}{HTML}{4C99FF}
\definecolor{mygreen}{HTML}{00CC66}

\definecolor{softgreen}{HTML}{E6FFF2} 
\definecolor{greenborder}{HTML}{00CC66}
\definecolor{visiblegreen}{HTML}{007F3F} %
\newcommand{\highlightcell}[1]{%
  \fcolorbox{greenborder}{softgreen}{\strut #1}%
}

\newcolumntype{Y}{>{\centering\arraybackslash}X}

\title{TRPrompt: Bootstrapping Query-Aware Prompt Optimization from Textual Rewards}

\author{%
  Andreea Nica \\
  EPFL \\
  \texttt{andreea.nica@epfl.ch} \\
  \And
  Ivan Zakazov\thanks{IZ and NMB co-supervised the project and contributed equally.} \\
  EPFL \\
  \texttt{ivan.zakazov@epfl.ch} \\
  \AND
  Nicolas Mario Baldwin\footnotemark[1] \\
  EPFL \\
  \texttt{nicolas.baldwin@epfl.ch} \\
  \And
  Saibo Geng \\
  EFPL \\
  \texttt{saibo.geng@epfl.ch} \\
  \And
  Robert West \\
  EPFL, Microsoft Research  \\
  \texttt{robert.west@epfl.ch} \\
}

\begin{document}

\maketitle

\begin{abstract}
  
Prompt optimization improves the reasoning abilities of large language models (LLMs) without requiring parameter updates to the \textit{target model}. Following heuristic-based \textit{"Think step by step"} approaches, the field has evolved in two main directions: while one group of methods uses textual feedback to elicit improved prompts from general-purpose LLMs in a training-free way, a concurrent line of research relies on numerical rewards to train a special \textit{prompt model}, tailored for providing optimal prompts to the \textit{target model}. In this paper, we introduce the \textbf{T}extual \textbf{R}eward \textbf{P}rompt framework (\textbf{TRPrompt}), which unifies these approaches by directly incorporating textual feedback into training of the \textit{prompt model}. Our framework does not require prior dataset collection and is being iteratively improved with the feedback on the generated prompts. When coupled with the capacity of an LLM to internalize the notion of what a "good" prompt is, the high-resolution signal provided by the textual rewards allows us to train a \textit{prompt model} yielding state-of-the-art query-specific prompts for the problems from the challenging math datasets \textit{GSMHard} and \textit{MATH}.
\end{abstract}

\section{Introduction}

The success of recent state-of-the-art large language models (LLMs) should not only be associated with massive scaling of parameters but also with aligning these models with human feedback \cite{ouyang2022training}. Reinforcement Learning from Human Feedback (RLHF) proved to be an extremely effective method to inject human knowledge into LLMs, enabling them to achieve human-level performance on a variety of benchmarks \cite{achiam2023gpt}. RLHF employs a reward model that generates \textit{numerical} rewards used to guide the optimisation of the language model. The numerical rewards act as proxy for human preference and play a crucial role in training. 
However, depending on the downstream task, numerical rewards can be sparse, uninformative and very difficult to define. Textual feedback, on the other hand, leverages the richness of language and can offer a more informative and nuanced learning signal. One such task where numerical rewards are hard to model is the \textbf{query-dependent prompt optimisation} problem, where the goal is to generate a prompt tailored to each input query, helping guide the language model toward high-quality outputs.

\begin{wrapfigure}{r}{0.51\textwidth}
  \centering
    \includegraphics[width=1\linewidth]{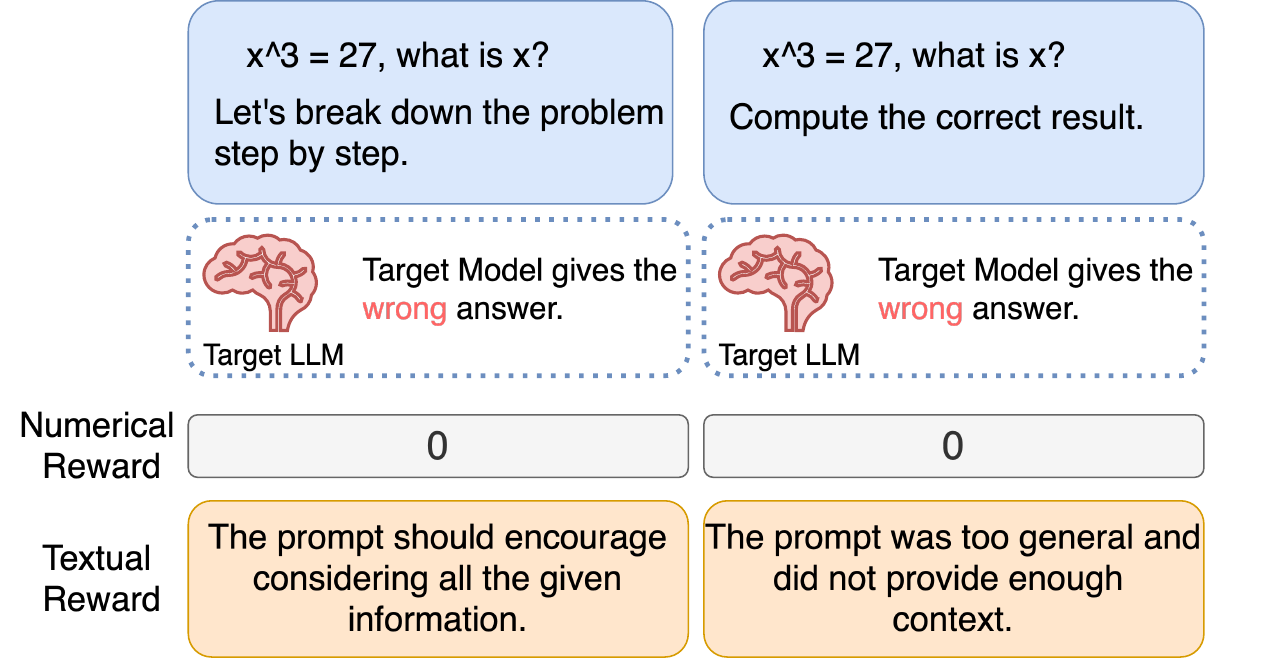}
    \caption{Motivating Example: Textual reward is more expressive than numerical reward. When using both prompts, the target model reaches a wrong answer. While the numerical reward is the same (0), the textual reward distinctively captures each prompt fails to guide the target language model in answering correctly.}
    \label{fig:motivation}
\end{wrapfigure}

Despite their remarkable power, LLMs seem to be challenged when it comes to mathematical and logical reasoning tasks \cite{li2024gsm}. Prompt engineering (concatenating the input query with an instruction) emerged as a lightweight solution to enhance the reasoning abilities of LLMs \cite{wei2022chain, kojima2022large} since it eliminates the need for parameter tuning. Research efforts have explored different methods of finding optimal prompts. One direction focuses on using LLMs as black-box optimizers to iteratively update the optimal prompt using textual feedback \cite{zhou2022large, liu2024large, pryzant2023automatic, yuksekgonul2024textgrad}. This train-free methods solely rely on the interactions with an off-the-shelf LLM. In contrast, other methods investigate training a policy model through reinforcement learning that is able to generate the optimal prompt \cite{sun2023query, deng2022rlprompt, li2024guiding, kong2024qpo}. Although most of these methods focus on finding an optimal task-level prompt \cite{deng2022rlprompt, zhou2022large} that achieves optimal average performance throughout the entire dataset, query-level prompt optimization techniques \cite{kong2024qpo, li2024guiding, sun2023query} can yield better results, since the prompt is specifically adapted to fit a particular query. Given the binary aspect of the query-dependent problem (the prompt guided the LLM to arrive at a correct/incorrect answer), designing a \textit{numerical reward} to successfully guide the prompt-space exploration for a given query is extremely challenging due to the difficulty in encoding nuanced preferences within sparse signals. Figure \ref{fig:motivation} shows a motivating example.

A \textit{textual reward} is an evaluation expressed in the natural language of a prompt, serving as a qualitative measure of its effectiveness in achieving a specific objective. Some successful applications of textual rewards focus on improving the quality of the reward models \cite{ye2024improving}, by concatenating synthetic critiques  with the inputs. The textual critiques improve the performance and data efficiency of the reward model. However, research on employing textual rewards for prompt optimization problems remains critically scarce.  

\textbf{Solution: TRPrompt.} To address the aforementioned limitations in the existing literature, in this paper, we propose TRPrompt, a Textual Reward Prompt optimization method that investigates the potential benefits of using textual rewards instead of numerical ones during the training of the prompt model. We believe that the richness of the textual reward is able to encapsulate more information about the performance of a given prompt compared to a numerical value. The textual reward should offer stronger signals and more robust features for fine-tuning the prompt model used to generate optimal prompts.

We develop a proof-of-concept framework that fine-tunes a relatively small model (Llama-3-8B-Instruct \cite{llama3modelcard}) to generate query-level optimal prompts. Since there is little work in this space, we have developed the entire methodology of applying textual rewards during training for prompt optimization problems, adapted for the query-dependent objective, where numerical rewards are harder to model. We train the prompt model through supervised fine-tuning (SFT) using a synthetically generated database of pairs of prompts and their corresponding textual rewards. The pipeline is fully automated, with textual rewards being generated by another language model that acts as a black-box reward model. In our experiments, we use a model within the same family (Llama-3-8B-Instruct) to generate textual rewards, facilitating the process of self-improvement. Consequently, we establish an iterative process where the fine-tuned prompt model efficiently generates improved, better prompts, which are then used to further fine-tune
the prompt model, creating a cyclical feedback loop. In this way, at each iteration, the prompt model receives signals through the textual rewards about the actual performance of the prompts that are considered optimal, facilitating improvement.

{
\setlength{\tabcolsep}{12pt} 

\begin{table}[h!]
\centering
\label{tab:only_one}
\caption{Overview of prompt optimization methods categorized by reward type and training setup. TRPrompt is the \textbf{only} method that combines textual rewards with a trainable approach.}
\begin{tabular}{@{} l | c c @{}}
\toprule
& \textbf{\textcolor{visiblegreen}{Train}} & \textbf{Train-free} \\
\midrule
\textbf{\textcolor{visiblegreen}{Textual Rewards}}   & \highlightcell{TRPrompt (ours)} & Textgrad \cite{yuksekgonul2024textgrad}, APO\cite{pryzant2023automatic} \\
\textbf{Numerical Rewards} & Prompt-OIRL \cite{sun2023query}, QPO\cite{kong2024qpo}           & APE\cite{zhou2022large} \\
\bottomrule
\end{tabular}

\end{table}
}

Our framework can be applied to a wide range of tasks for evaluating the mathematical reasoning skills of LLMs. TRPrompt can be initialized on any off-the-shelf LLM and does not require prior dataset collection, which makes it instantly adaptable to any dataset.  We perform experiments on multiple datasets and analyze the effectiveness of our approach in producing optimal prompts for the desired objective. Our contributions can be summarized as follows:
\begin{itemize}
    \item We advance the concept of textual rewards by being the first to incorporate them as a supervision signal directly used during training — moving beyond prior train-free approaches such as TextGrad\cite{yuksekgonul2024textgrad}.
    \item We propose and develop a novel methodology TRPrompt that is able to exploit the benefits of textual rewards for the query-dependent prompt optimisation objective to improve prompting performance. Table \ref{tab:only_one} compares prior methods with ours, showing how TRPrompt is the first to embed textual rewards directly into the training loop.
    \item We validate the efficacy of our method in various data sets, providing a comprehensive analysis of its performance.
    \item We achieve state-of-the-art performance in prompt optimization on the more challenging datasets, GSMHard and MATH, demonstrating the effectiveness of our method.
\end{itemize}

\begin{figure}[htbp]
    \centering
    \includegraphics[width=0.75\linewidth]{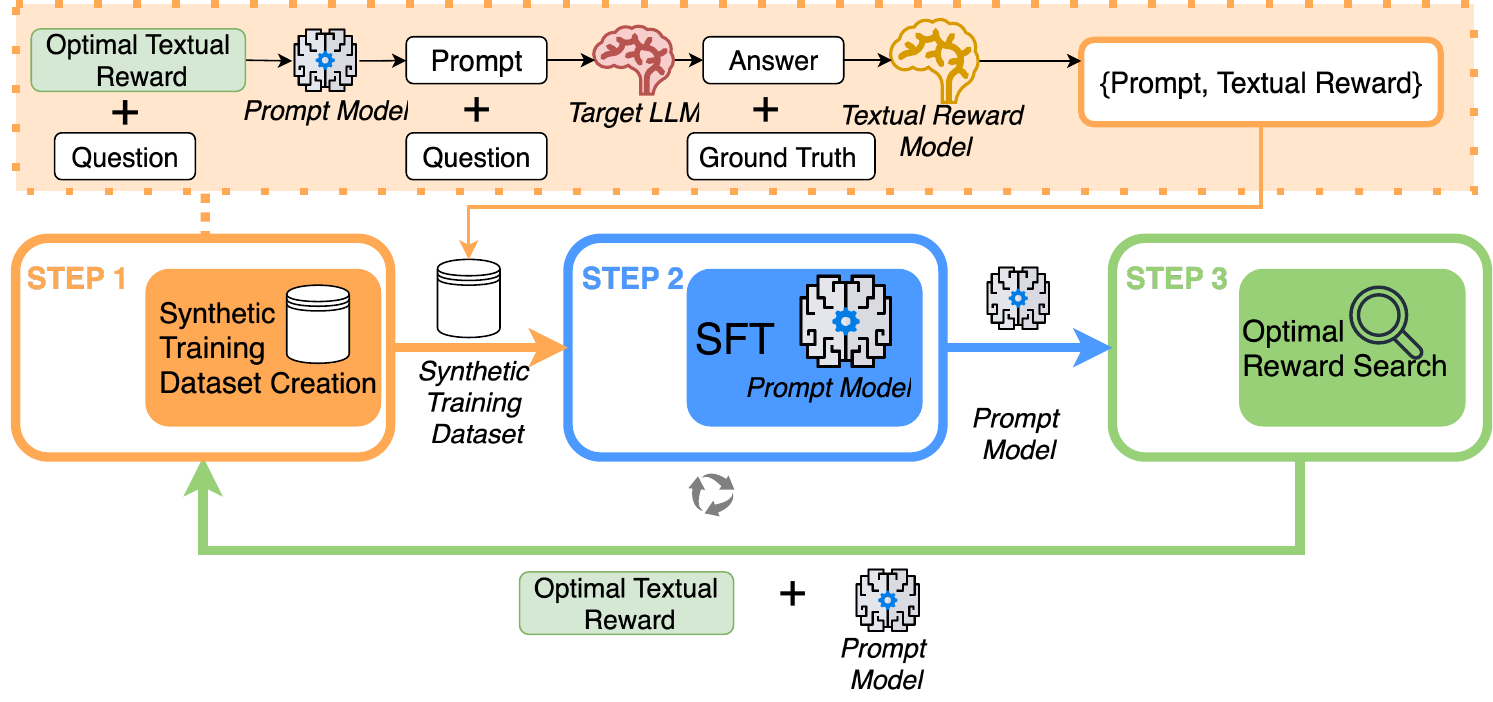}
    \caption{Overview of our query-dependent prompt optimization iterative pipeline, containing 3 steps: (1) synthetic training dataset creation, (2) fine-tuning of the prompt model and (3) optimal reward search.}
    \label{fig:query_overall}
\end{figure}

\section{The query-dependent prompting problem}
\begin{wrapfigure}{r}{0.27\textwidth}
  \centering
    \includegraphics[width=1\linewidth]{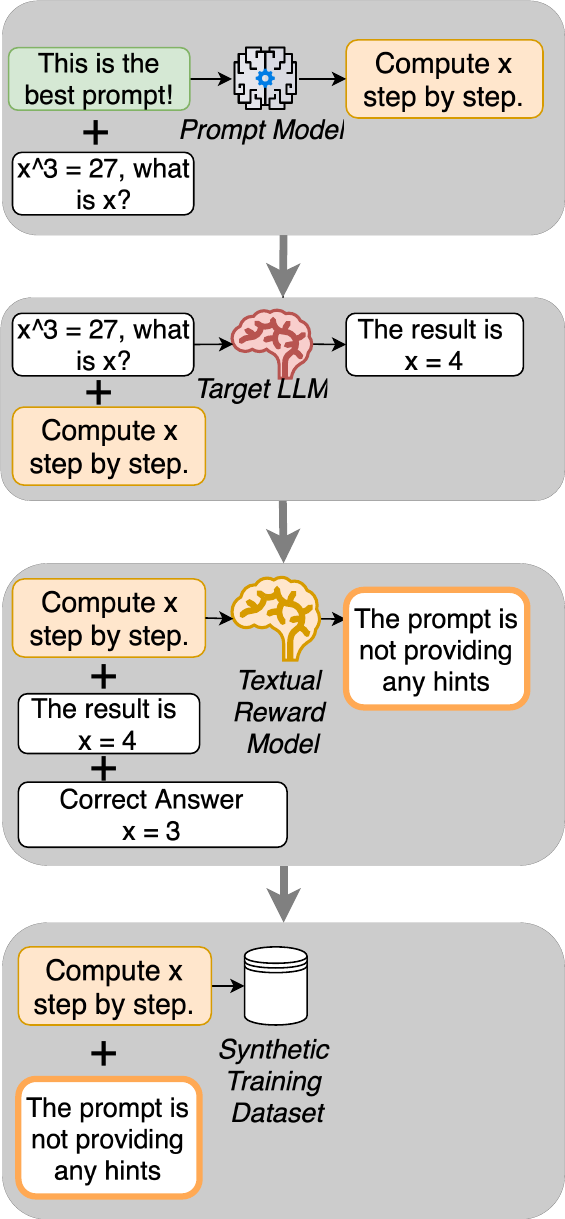}
    \caption{\textcolor{myorange}{\textbf{Step 1:}} Prompt generation and textual reward calculation}
    \label{fig:example_gen}
    \vspace{-2.5em}
\end{wrapfigure}

The query-dependent prompting problem centers on guiding a target model to correctly solve a given question. In this work, we focus specifically on mathematical questions. To achieve this, we prepend the question with a query-dependent prompt -- an instruction specifically tailored to the given question to improve the model’s reasoning. This prompt is generated by a separate prompt model, which is trained to produce effective query-dependent instructions. The prompt model conditions its generations on textual rewards — natural language feedback that indicates the quality of the desired prompt. To generate prompts that reliably lead the target model to correct answers, we condition the prompt model on the optimal textual reward, i.e., the best feedback that a prompt can receive for a given question. This process is illustrated in Figure \ref{fig:query_overall}, Step 1. A concrete example is shown in Figure \ref{fig:example_gen}. Thus, training the prompt model involves two interconnected goals: (1) fine-tuning the prompt model  and (2) identifying the optimal textual reward. This process constitutes what we term the query-dependent prompting problem, which we now formalize.

\textbf{Questions and answers.} We consider the task of answering mathematical questions \(q \in \mathcal{Q} = \mathcal{V}^{\infty}\) where each question is expressed in natural language using vocabulary \(\mathcal{V}\). Each question \(q\) is expected to have a mathematically correct answer \(y^{*} \in \mathcal{Y}_{gt} \subset \mathcal{Y}\) which will be considered the ground-truth. 

\textbf{Prompting.} The performance of a target LLM in answering questions can be enhanced by using appropriate prompting. A prompt \(p \in \mathcal{P} = \in \mathcal{V}^{\infty}\) is a combination of words in the vocabulary \(\mathcal{V}\), an instruction in natural language that explains how to solve a task. In this work, prompts are appended at the end of a question. A \textbf{query-dependent prompt} is a prompt generated specifically to guide the target model to the correct result for a specific question.

\textbf{Target LLM.} The mathematical questions are answered using a target LLM  \( M_{\text{target}} \colon \mathcal{Q} \to \mathcal{Y}\) to obtain answers. The generated answers can be mathematically described as \( \hat{y}_{i} = M_{\text{target}}(q_{i}, p_{i}) \) where \(q_{i}\) is the \(i\)-th question and \(p_{i}\) is the query-dependent prompt corresponding to question \(q_{i}\) . 

The quality of these generations can be evaluated using the following metric \(r(y^{*}_{i}, \hat{y}_{i}) = \mathbbm{1}\{ \hat{y}_{i} = y^{*}_{i} \}\), which checks for equality between the generated answer and the ground-truth.

\textbf{Textual reward.} In order to assess the quality of a query-dependent prompt, we will use a textual reward \(t \in \mathcal{T} = \mathcal{V}^{\infty}\). The textual reward represents feedback expressed in natural language form that will synthesize the performance of the query-dependent prompt in guiding the Target LLM to the correct answer.

\textbf{Optimal textual reward.} We define \(t^* \in \mathcal{T}\) as the optimal textual reward, corresponding to the best achievable feedback that a query-dependent prompt can obtain, in line with the objective scope.

\textbf{Textual reward model.} A textual reward model \(R_{\text{textual}}\) where \(R_{\text{textual}}  \colon P \times \mathcal{Q} \times \mathcal{Y}_{\text{gen}} \times \mathcal{Y}_{\text{gt}} \to \mathcal{T} \) is a language model that can produce textual reward \(t \in \mathcal{T}\) for a query-dependent prompt \(p_{i} \in \mathcal{P}\) based on the answer generated by the Target LLM \(\hat{y}_{i} \in \mathcal{Y}_{\text{gen}} \subset \mathcal{Y}\) and the ground-truth for the question \(y^{*}_{i} \in \mathcal{Y}_{\text{gt}} \subset \mathcal{Y}\). We have \(t = R_{\text{textual}}(q_{i}, p_{i}, \hat{y}_{i}, y^{*}_{i})\).

\textbf{Prompt model.} A prompt model \(\Pi_{\text{query}}: \mathcal{Q} \times \mathcal{T} \to P\)  is a language model that is used to generate query-dependent prompts. For some query \(q_{i} \in \mathcal{Q}\) and a textual feedback \(t \in \mathcal{T}\),  we have \(\Pi_{query}(q_{i}, t) = p_{i} \in \mathcal{P}\). The role of the prompt model is to produce a query-dependent prompt, instruction that will enhance the performance of the target model in the way specified by the feedback \(t\) for the specific question \(q_{i}\). 

\textbf{Dual objective.} To maximize the performance of the Target LLM on reasoning tasks, the query-dependent prompting problem can be formulated as a dual-objective optimization problem.
\begin{enumerate}
    \item \textbf{Prompt model objective. }The first objective is to optimize the prompt model \(\Pi_{\text{query}}\) to generate query-dependent prompts corresponding to the optimal textual reward \(t^*\) for the given questions. These prompt should guide the Target LLM towards producing the correct answers. The goal is to maximize the target model’s performance by ensuring the prompts satisfies the highest-quality feedback signal.
    \begin{equation}
    \Pi^*_{\text{query}} = \arg\max_{\Pi_{\text{query}}} r(y^{*}_{i}, \, M_{\text{target}}( q_{i}, \Pi_{\text{query}}(q_{i}, t^{*})))_{i \in [N], N=|\mathcal{Q}|}
    \label{eq:query_problem}
    \end{equation}
    
    \item \textbf{Optimal textual reward objective.} The second objective focuses on identifying the optimal textual reward \(t^{*}\) for a given prompt model \(\Pi_{query}\). This reward represents the most informative and effective natural language feedback that, when used to condition the prompt model, results in query-dependent prompts that maximizes the target model's accuracy.
    \begin{equation}
    t^{*} = \arg\max_{t \in \mathcal{T}} r(y^{*}_{i}, M_{\text{target}}(q_{i}, \Pi_{\text{query}}(q_{i}, t)))_{i \in [N], N=|\mathcal{Q}|}
    \label{eq:optimum_objective}
    \end{equation}
\end{enumerate}

\section{Prompt optimization with textual reward} \label{method}

Existing query-dependent methods rely exclusively on numerical rewards to guide training \cite{sun2023query, kong2024qpo}. In contrast, we argue that textual rewards provide a richer and more expressive supervisory signal, capable of more effectively steering the learning process.
We introduce \textbf{TRPrompt}, a framework for prompt optimization that leverages textual rewards as the primary training signal. Our iterative, multi-loop design enables continuous improvement, with textual feedback facilitating effective knowledge transfer to the prompt model throughout the learning process.

 Figure \ref{fig:query_overall} displays our iterative approach on query-dependent optimization problem. The pipeline contains three important steps: \textcolor{myorange}{\textbf{Step 1.}} synthetic training dataset creation, \textbf{\textcolor{myblue}{Step 2.}} prompt model fine-tuning and \textcolor{mygreen}{\textbf{Step 3.}} optimal reward update. 

 \textbf{Step 1. Query-dependent prompt generation  \& textual reward calculation.} The synthetic training dataset contains query-dependent prompts and their textual rewards. To generate the question-specific prompt, an optimal feedback signal \(t^* \in \mathcal{T}\) is provided together with the question \(q \in \mathcal{Q}\) to the prompt model \(\Pi_{\text{query}}\):
\[
p_i = \Pi_{\text{query}}(q_i, t^*) \in \mathcal{P}, \quad \forall q_i \in \mathcal{Q}
\]
The query-specific generated prompt is then concatenated with the question to produce the answer from the Target LLM. 
\[
y_i = M_{\text{target}}(q_i,\, p_i), \quad \forall q_i \in \mathcal{Q}, \,p_i = \Pi_{\text{query}}(q_i, t^*)
\]

Analysing the generated answer and the ground-truth, the  textual reward model \(R_{\text{textual}}  \colon P \times \mathcal{Q} \times \mathcal{Y_{\text{gen}}} \times \mathcal{Y_{\text{gt}}} \to \mathcal{T}\) produces a detailed textual feedback regarding the performance of the query-dependent prompt for the specific task. 
\[
R_{\text{textual}}(p_i,\, (q_i, y_i, y^*_i)) = t_i
\]
The textual rewards corresponding to a prompt for a specific task are collected for fine-tuning \(\mathcal{D_{\text{train}}} = \{(p_i, q_i, t_i)\}\). A concrete example of the entire flow of STEP 1 can be seen in Figure \ref{fig:example_gen}

 \textbf{Step 2. Prompt model fine-tuning.}
For the query-specific task, we fine-tune the prompt model \(\Pi_{\text{query}}\) on the prompts \(p \in \mathcal{P}\) conditioned on the textual reward \(t \in \mathcal{T}\) and the question \(q \in \mathcal{Q}\). Specifically, we maximize the log-likelihood:
\begin{equation}
    \mathcal{L}_{SFT} = -\mathbb{E}_{(p_i, q_i, t_i)\sim\mathcal{D_{\text{train}}}} \log \mathbb{P}(p_i|q_i, t_i)
\end{equation}

\textbf{Step 3. Optimal textual reward update.} Since the optimal reward is a static value that has not been seen during training, finding the best possible optimal reward that will leverage all the knowledge accumulated throughout fine-tuning is an optimization problem on its own. Specifically, we want to find a textual reward inline with the objective from Equation \ref{eq:optimum_objective}. To search for the best optimal textual reward, we employ a train-free optimization strategy, Textgrad. \cite{yuksekgonul2024textgrad}. Using Textgrad, we are able to explore the reward space. This step is completed after each round of fine-tuning, to get the corresponding optimal reward for each model. At the end of fine-tuning the prompt model \(\Pi_{\text{query}}\) we have: 
\[
t^* = Textgrad( \Pi_{\text{query}})
\]

The iterative procedure for the query-dependent objective, repeating Steps 1, 2, and 3 over \(K\) iterations, is detailed in Algorithm \ref{alg:query_dependent}.

\begin{algorithm}
\caption{Query-Dependent Prompt Optimization}
\label{alg:query_dependent}
{\textbf{Require:} Question dataset \(\mathcal{Q}\), number of iterations \(K\), a language model as textual reward model \(R_{\text{textual}}\), a Target LLM \(M_{\text{target}}\), SFT(\(\cdot\)) denotes the supervised fine-tuning process, Textgrad(\(\cdot\)) denotes applying Textgrad for optimal reward search, an initial optimal textual reward \(t^*_0\), the number of questions to analyze the textual reward \(N\), a Prompt Mode \(\Pi_{\text{query}}\) that generates query-dependent prompts.}
\begin{algorithmic}[1]
\For{\(k = 1\) to \(K\)}
    \State \(\mathcal{D}_{\text{train}} \leftarrow \emptyset\)
    \For{\(q_i\) in \(\mathcal{Q}\)}
        \State \textbf{Prompt Generation:} \(p_i \leftarrow \Pi^{k-1}_{\text{query}}(q_i, t^*_{k-1})\)
        \State \textbf{Question Answering:} \(y_i = M_{\text{target}}(q_i,\, p_i)\)
        \State \textbf{Textual Reward} \(t_i \leftarrow R_{\text{textual}}(p_i, (q_i, y_i, y^*_i)\)
        \State \(\mathcal{D}_{\text{train}} \leftarrow \mathcal{D}_{\text{train}} \cup \{(p_i, \, q_i, \,t_i)\}\)
    \EndFor
    \State \textbf{Fine-tuning: }\(\Pi^k_{\text{query}} \leftarrow SFT(\Pi^{k - 1}_{\text{query}}, \, \mathcal{D}_{\text{train}})\)
    \State \textbf{Optimal Reward Update}: \(t^*_k = Textgrad( \Pi^k_{\text{query}}, t^*_{k-1})\)
    \EndFor
\State \Return  Prompt Model \(\Pi^K_{\text{query}}\) and optimal reward \(t^*_K\) which can generate optimal prompts for a specific task.
\end{algorithmic}
\end{algorithm}

\section{Related work}

Prompt optimization has followed two main paths: training-based methods using numerical rewards \cite{sun2023query, kong2024qpo}, and train-free approaches leveraging natural language to refine prompt outputs without weight updates \cite{yuksekgonul2024textgrad,pryzant2023automatic}. TRPrompt bridges these paradigms by enabling gradient-based training directly from textual rewards.

\textbf{Prompt optimisation.} Discrete prompts have one important feature: interpretability. Research on zero-shot prompting has shown that LLMs are good zero-shot reasoners simply by using the prompt 'Let's think step by step' (CoT) \cite{kojima2022large}. Although most prompt-engineering research efforts have focused on finding the optimal prompts at the task-level (a prompt efficient for the entire dataset), this approach might yield suboptimal results at the per-instance level. A generally effective prompt, while performing well on average, may steer the LLM towards incorrect results \cite{sun2023query}. This is why increasing attention is now being placed on query-dependent methods  \cite{sun2023query, zhang2022tempera,li2024guiding,kong2024qpo}. Prompt-OIRL \cite{sun2023query} uses inverse RL to train a reward model which selects the best one of the n-candidate prompts for a query. QPO \cite{kong2024qpo} uses reinforcement learning guided by a custom numerical reward to fine-tune a small pre-trained language model to generate query-dependent optimal prompts tailored to the input queries. All of the aforementioned methods uniquely rely on numerical rewards to guide the training process.

\textbf{Finetuning conditioned on reward.} Conditioning generation on reward signals during supervised learning has been explored in prior work \cite{Zhang23, shypula2023learning}, showing success in domains such as code optimization\cite{shypula2023learning}. However, these methods rely solely on numerical reward labels and do not leverage the richer supervision offered by textual rewards.

\textbf{Optimization with natural language feedback.} Recent lines of work investigate the use of natural language feedback to improve the LLM's overall performance. One line of research utilizes textual feedback for self-improvement by refining the model outputs into better ones. These applications deploy LLMs as black-box optimizers in a train-free method where the textual feedback is used to obtain iterative improvement in model's output \cite{pryzant2023automatic,yuksekgonul2024textgrad}. Textgrad \cite{yuksekgonul2024textgrad} takes inspiration from gradient descent, providing a customizable framework that can apply a "textual gradient" (textual feedback)  to improve the generation of an LLM on a variety of tasks, including prompt optimization. TPO \cite{li2025test} translates a numerical reward signal generated by the reward model into textual critiques and uses them as textual rewards to iteratively refine the final output of the model. Textual rewards are used to help the policy model come with suggestions guiding the generations for the next iteration. Another line of research explores using critiques for training reward models \cite{ye2024improving}, thus converting the richer signals of the synthetic critiques into a numerical representation. However, no method to our knowledge uses textual rewards as direct signals during training.

\begin{table}[h!]
\centering
\label{tab:differences}
\caption{TRPrompt  differentiates from existing research by (1) not requiring a set of manually defined prompts, (2) being completely independent from the starting prompts since our method constructs the entire training dataset synthetically, avoiding the need for initial prompts, (3) offering a query-dependent methodology for the prompt optimisation problem, (4) relying on textual rewards that are used as the main signal during training (5).}
\vspace{0.5em}

\begin{tabularx}{\textwidth}{@{}l|YYYYY@{}}
\toprule
\textbf{Method} & \textbf{(1) Avoids Static/Manual Prompts} & \textbf{(2) Independent of Initial Prompts} & \textbf{(3) Query Dependent Prompt} & \textbf{(4) Textual Rewards} & \textbf{(5) Textual Rewards For Training} \\
\midrule
CoT\cite{kojima2022large} & \xmark & \xmark & \xmark & \xmark & \xmark \\
Prompt-OIRL \cite{sun2023query}  & \xmark & \xmark & \cmark & \xmark & \xmark \\
QPO \cite{kong2024qpo}  & \cmark & \xmark & \cmark & \xmark & \xmark \\
Textgrad \cite{yuksekgonul2024textgrad} & \cmark & \xmark & \xmark & \cmark & \xmark \\
TPO \cite{li2025test} & \xmark & \xmark & \xmark & \cmark & \xmark \\
\midrule
\textbf{TRPrompt (ours)} & \cmark & \cmark & \cmark & \cmark & \cmark \\
\bottomrule
\end{tabularx}

\vspace{0.5em}

\end{table}

\section{Experiments} \label{experiment}
In this section, we present empirical evidence illustrating the efficacy of our method using textual rewards in guiding the training of the optimal query-dependent prompt model. We start by outlining the general
experimental setup, followed by a in-depth analysis of results. 

\textbf{Tasks.} We perform experiments on 3 math-reasoning datasets of various complexities: GSM8K \cite{cobbe2021training}, GSMHard \cite{gao2022pal} which is a harder version of GSM8K and MATH \cite{hendrycksmath2021}. These arithmetic reasoning tasks are widely studied in the zero-shot prompting domain.

\textbf{Models.} Our framework consists of three distinct models: \textbf{1. Target LLM} for which the optimal prompts are being searched.  \textbf{2. Textual reward model} that generates the textual rewards. \textbf{3. Prompt model} that will generate the optimal prompts for the Target LLM, the model being fine-tuned. We set all these models to be Meta-Llama-3-8B-Instruct \cite{llama3modelcard}. Our method is model-agnostic so different model instances can be changed for different components. However, we choose to explore a same-model approach since we want to investigate the ability of the model for self-improvement. For Textgrad\cite{yuksekgonul2024textgrad}, we used GPT-4o-mini to do the optimal textual reward  updates. 
\textbf{Technical details} about the experiments can be found in the Technical Appendix.

\begin{table}[h!]
\centering
\caption{Comparison between state-of-the-art methods and TRPrompt on GSM8K, GSMHard, and MATH datasets.}
\label{tab:all_results}
\vspace{0.5em}

\begin{tabular}{lccc}
\toprule
\textbf{Name} & \textbf{GSM8K (Acc.)} & \textbf{GSMHard (Acc.)} & \textbf{MATH (Acc.)} \\
\midrule
CoT \cite{kojima2022large} & 85.59\% & 27.98\% & 39.35\% \\
Prompt-OIRL (6 prompts) \cite{sun2023query} & 84.53\% & 28.61\% & 21.31\% \\
QPO (1 starting prompt) \cite{kong2024qpo} & 84.76\% & 27.98\% & 28.37\% \\
QPO (500 starting prompts) \cite{kong2024qpo} & \textbf{86.05\%} & 30.80\% & 37.31\% \\

\midrule
\textbf{TRPrompt Query-Dependent (Ours)} & 84.53\% & \textbf{31.76\%} & \textbf{41.37\%} \\
\bottomrule
\end{tabular}
\end{table}

Table \ref{tab:all_results} presents a comprehensive comparison between the query-dependent optimized prompts generated by TRPrompt against several SOTA methods: CoT "Let's think step by step" prompt, query-dependent prompts produced by QPO under two distinct initialization configurations (initial dataset containing 1 expert prompt vs 500 expert prompts), Prompt-OIRL with reward learned on 6 expert prompts and the task-level prompt obtained by applying Textgrad.

From the results, we can conclude: \textbf{(1)}. On challenging mathematical benchmarks such as GSMHard and MATH, where the target model demonstrates difficulty in generating correct answers, TRPrompt outperforms all other state-of-the-art methods by +1\% on GSMHard and by +2\% on MATH. In the case of GSM8K, where the model answers most questions correctly by default, textual rewards appear to provide a weaker signal compared to their effectiveness on more challenging datasets like GSMHard and MATH. \textbf{(2).} While QPO and Prompt-OIRL exhibit a strong dependence on initial expert-provided prompts, TRPrompt avoids this limitation entirely by eliminating the need
for handcrafted starting prompts, thereby avoiding initialization bias. Prompt-OIRL selects one of the predefined prompts, whereas QPO generates new prompts that are heavily influenced by the structure and content of the initial prompt set. The accuracy improves as the number of initial prompts is increased.
TRPrompt explores the prompt space more freely, resulting in higher-quality query-dependent prompts.

\begin{figure}[ht]
    \centering
    \begin{minipage}{0.34\textwidth}
        \centering
          \includegraphics[width=0.95\linewidth]{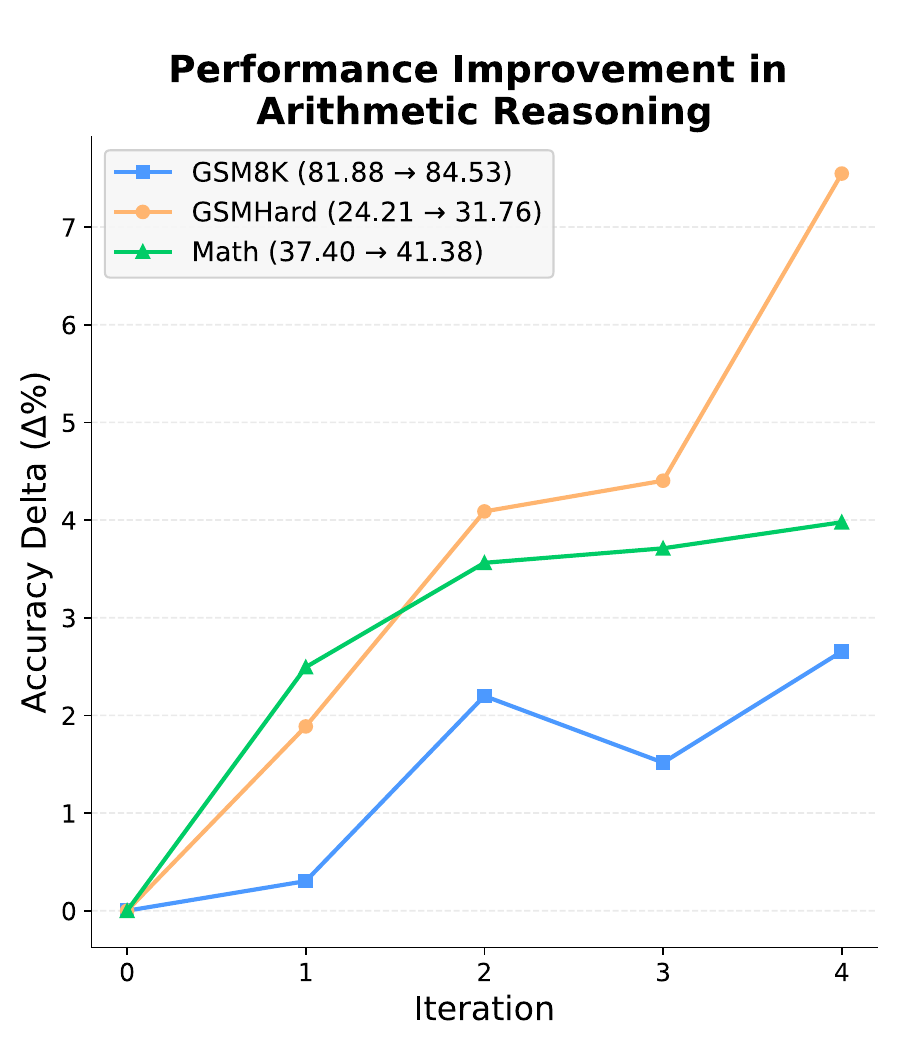}
          \caption{Accuracy gain (in percentage points difference) compared to the base model across iterations. Each iteration improves the ability of the prompt model to generate prompts that guide the target model to the correct answer.}
          \label{fig:iter}
        
    \end{minipage}%
    \hfill
    \begin{minipage}{0.63\textwidth}
        \centering
    \includegraphics[width=0.95\linewidth]{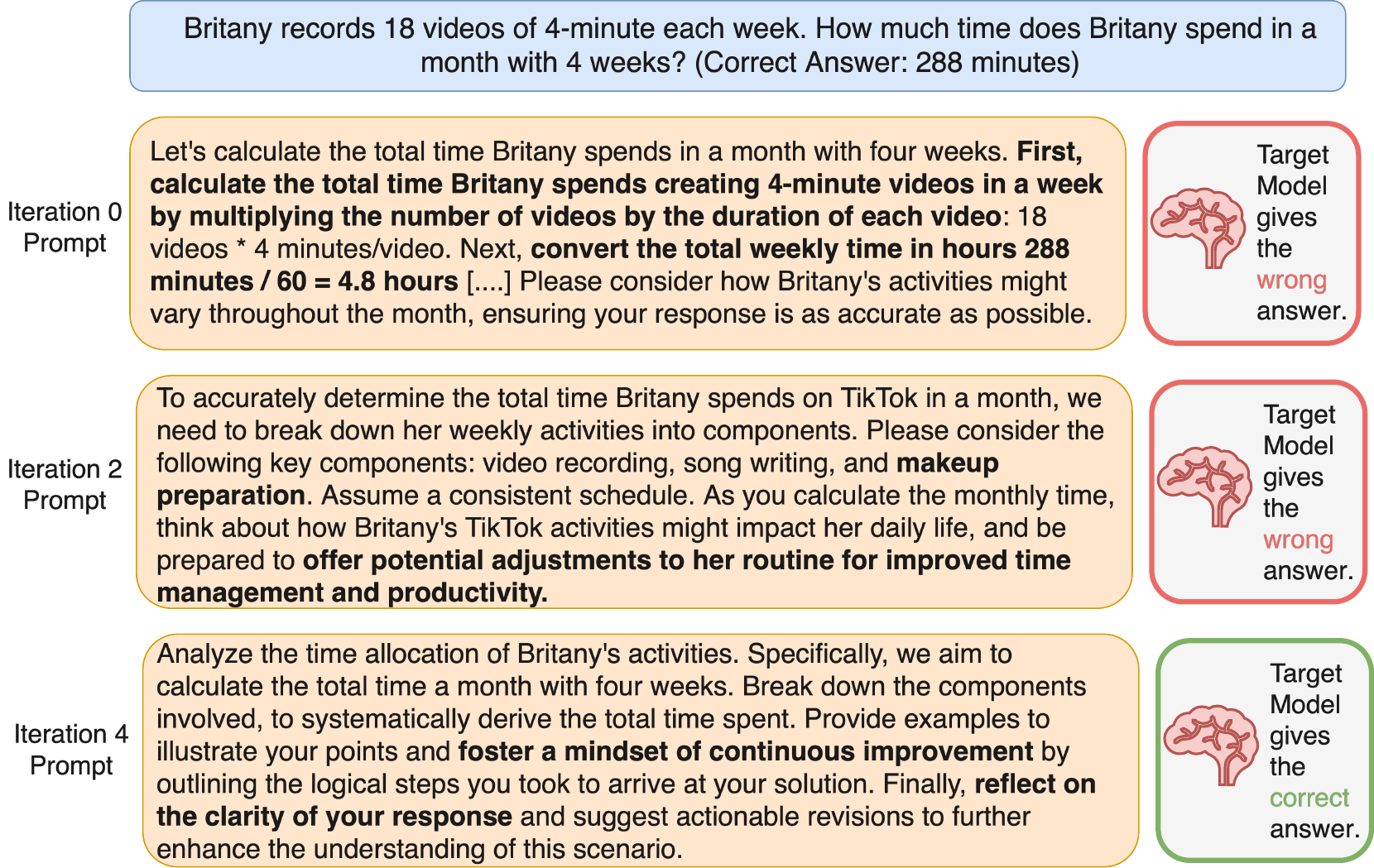}
    \caption{\textbf{Iterative improvement of query-dependent prompts on a GSMHard question.} Prompts generated by the prompt model become increasingly structured and general across iterations, reducing reliance on partial solutions and irrelevant details. By the final iteration, the improved prompt guides the target model to the correct answer.}
    \label{fig:improvement}    
    \end{minipage}
\end{figure}

\subsection{Analysis} \label{analysis}

\textbf{Iteratively better prompts.} The iterative nature of our pipeline is essential for ensuring that the prompt model learns from its own limitations. For each iteration, a new synthetic training dataset is constructed using the prompts generated by the most recent version of the prompt model, conditioned on the current optimal textual reward.
This setup guarantees that the textual reward is addressing the model's latest behavior, allowing it to receive precise critiques. 
As a result, the prompt model progressively refines its generations by learning from past errors, enabling targeted improvements. Figure \ref{fig:iter} shows the increasing improvement in prompt quality across iterations, with the fine-tuned prompt model achieving better performance ( + 7.5\%) than the base model over the course of training, especially for the harder datasets GSMHard and MATH. Each iteration exploits newly learned patterns through the textual rewards, gradually improving the quality of the generated prompts. Figure \ref{fig:improvement} exemplifies the progress of a prompt across iterations for a GSMHard question. The prompt model transitions from including explicit computations, to producing generic and occasionally irrelevant suggestions, before converging on effective, task-aligned instructions.

For the harder datasets (MATH and GSMHard), the target model struggles to generate correct answers, allowing textual rewards to provide highly targeted and constructive feedback. As a result, the prompt model shows substantial improvement from one iteration to the next, effectively learning from its errors. In contrast, on simpler datasets like GSM8K —where the target model already performs well without any prompt — the corrective power of textual rewards is diminished, leading to more modest gains. We hypothesize that the limited performance gains on GSM8K are partly due to an unbalanced training set dominated by positive feedback. When most prompts already lead to correct answers, the textual rewards overwhelmingly reinforce success, providing little contrast between effective and ineffective prompts. The lack of negative supervision undermines the quality of training, as the prompt model receives insufficient signal to refine its behavior.

\begin{figure}[ht]
    \centering
    \begin{minipage}{0.48\textwidth}
        \centering
        \includegraphics[width=0.72\linewidth]{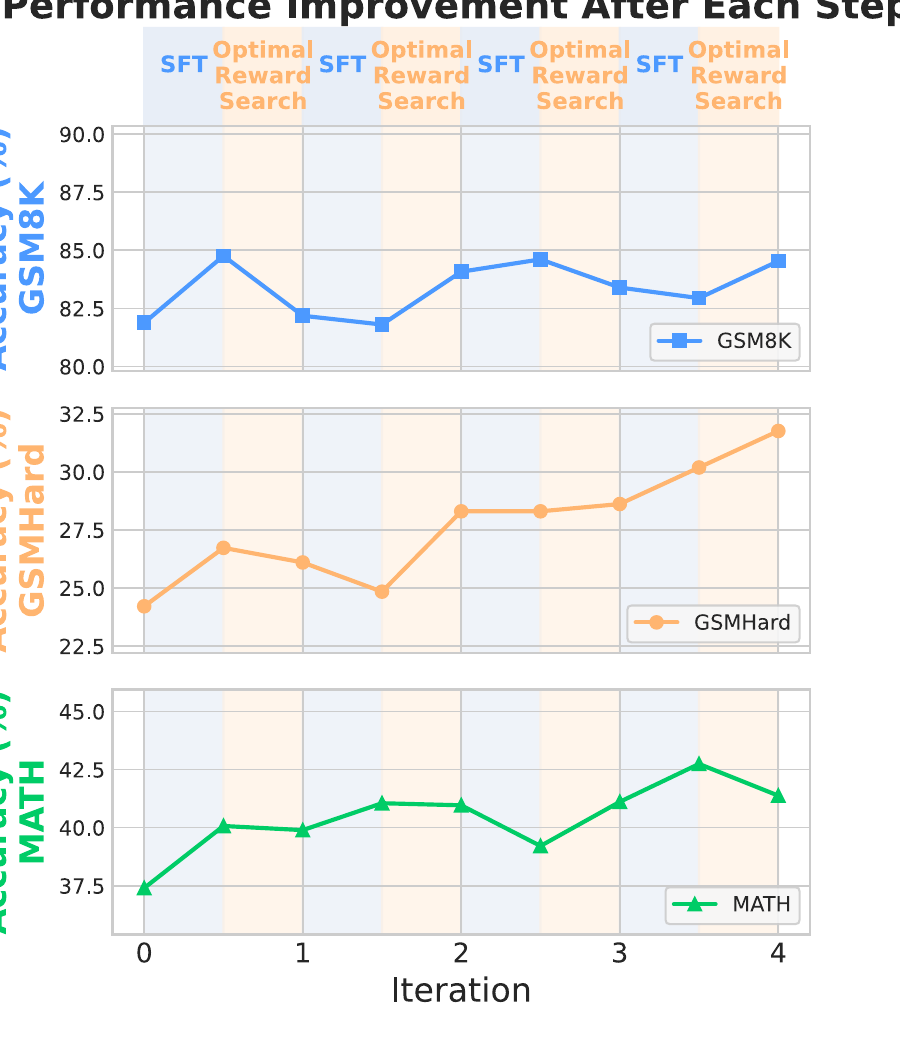}
        \caption{Intermediate performance obtained by the prompt model after each step in our pipeline. Both SFT and the Optimal Reward Search step contribute to enhancing the prompt model ability to generate efficient prompts.}
        \label{fig:ablation_steps}
        
    \end{minipage}%
    \hfill
    \begin{minipage}{0.48\textwidth}
        \centering
        \includegraphics[width=0.72\linewidth]{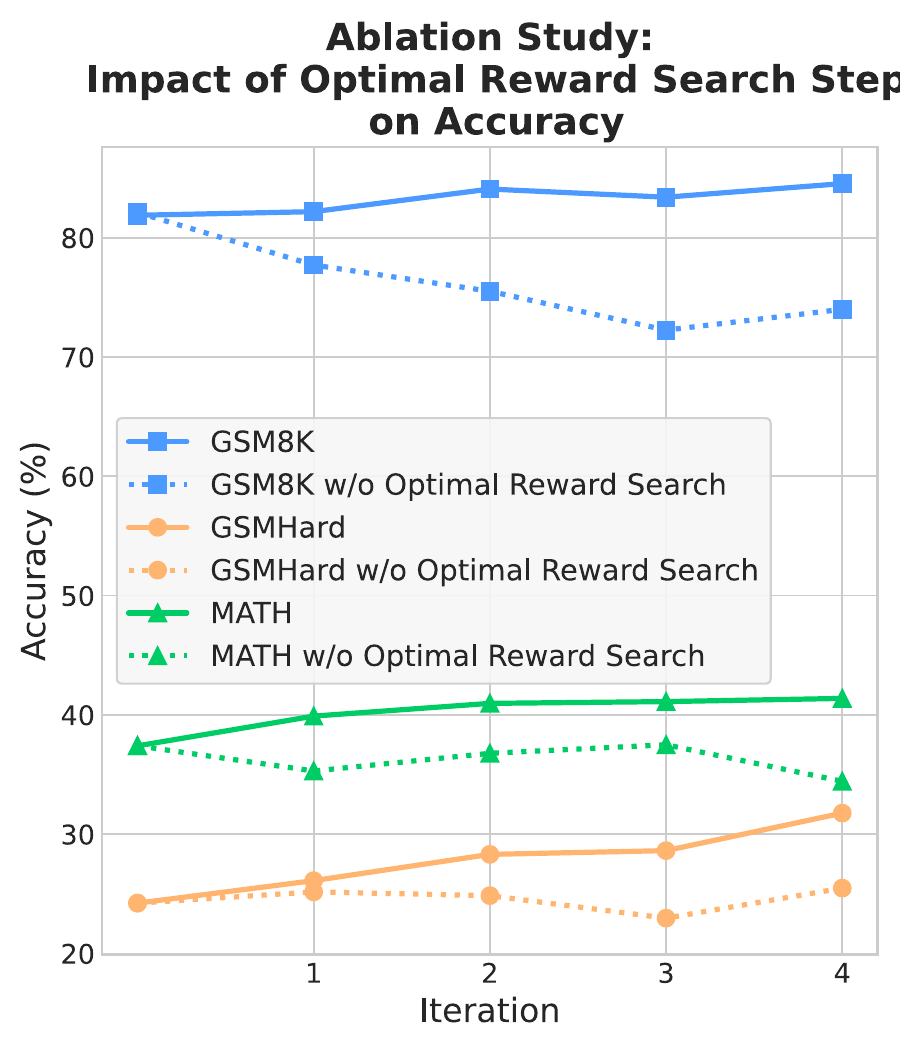}
        \caption{Impact of eliminating the Optimal Reward Search step from our pipeline: accuracy is consistently decreasing across iterations.}
        \label{fig:ablation_textgrad}     
    \end{minipage}
\end{figure}

\textbf{Ablation insights: why SFT + optimal reward search matter.} We investigate the impact of each step on the efficiency of our method. We compute the accuracy after each intermediate step in the pipeline and report the results for each dataset. Figure \ref{fig:ablation_steps} illustrates the impact of each step in the consistent increase in accuracy across iterations as the prompt model is fine-tuned using Supervised Fine-Tuning (SFT) and then applying Optimal Reward Search. The performance gains result from the combined effect of both steps, jointly enabling effective exploration of the prompt space. In contrast, Figure \ref{fig:ablation_textgrad} presents an ablation where the Optimal Reward Search step is removed from our framework and only SFT is applied. In this setting, performance gradually declines with each iteration across all datasets, suggesting that without updating the optimal textual reward, the prompt model begins to overfit to suboptimal feedback. These results highlight the critical role of Optimal Reward Search in guiding the prompt model towards meaningful exploration of the prompt space and sustaining performance improvements over time.

\textbf{Evaluating cross-dataset transfer: accuracy of prompt models on unseen tasks.} 
We evaluate the generalization ability of TRPrompt and query-dependent baselines by testing whether a prompt model trained on one benchmark can generate effective prompts for unseen datasets. For this experiment, the prompt model is trained on a single dataset and evaluated on the remaining datasets. As shown in Table \ref{tab:cross_generalization}, all methods show a drop in performance when applied out-of-distribution, confirming that prompt optimization is sensitive to the training distribution. However, TRPrompt achieves the highest accuracy on the MATH dataset when the prompt model is trained on either GSM8K or GSMHard, consistently outperforming Prompt-OIRL and QPO. Although TRPrompt does not exceed baseline performance on GSM8K when trained on GSMHard, this observation aligns with prior findings suggesting that performance gains are typically less pronounced on simpler datasets. Collectively, these results highlight the superior generalization capabilities of TRPrompt, particularly in more challenging domains. 

\begin{table}[h!]
\centering
\caption{ Cross-dataset generalization: accuracy when models trained on one dataset are evaluated on another datasets.}
\label{tab:cross_generalization}
\vspace{0.5em}

\begin{tabular}{llcccc}
\toprule
\textbf{Trained on ↓} & \textbf{Method} & \textbf{Tested on →} & \textbf{GSM8K} & \textbf{GSMHard} & \textbf{MATH} \\

               &                     &  & Accuracy       & Accuracy         & Accuracy \\
\midrule
GSM8K & Prompt-OIRL   & & -  & \textbf{29.55}\% & 22.14\% \\
         
  & QPO & & - & 28.61\% & 36.36\% \\
  & \textbf{TRPrompt (Ours)} & & - & 27.67\% & \textbf{37.20}\% \\ 
\midrule
GSMHard & Prompt-OIRL    & & \textbf{84.15}\% & -  & 26.53\% \\
        & QPO & &  82.94\% & - & 26.35\% \\
        & \textbf{TRPrompt (Ours)} &  & 83.09\% & - & \textbf{39.41}\% \\
                        
\bottomrule
\end{tabular}
\end{table}

\section{Limitations \& future work} \label{limitation}
\textbf{Reduced gains on easier datasets.} TRPrompt shows relatively low performance gains on GSM8K, both in-distribution and in out-of-distribution settings. The dominance of correct answer and positive textual rewards weakens the learning signal. Addressing this issue may require rebalancing the training data or introducing more nuanced textual feedback to better reflect subtle variations in prompt quality.

\textbf{High computational cost.} The Optimal Reward Search step involving Textgrad is computationally expensive and difficult to parallelize, making it the main bottleneck in the pipeline. This limits scalability and slows down training, especially on larger datasets. Future work could focus on more efficient optimal reward search methods.

\textbf{Further leveraging textual rewards.} Our framework can be extended to tasks where numerical rewards are difficult or unnatural to define, such as creative writing or poetry. In these domains, quantitative metrics often fail to capture quality, coherence, or style, whereas textual feedback can provide richer, more interpretable signals. TRPrompt offers a promising foundation for future research in this area.

\section{Conclusion}
We introduce TRPrompt, a novel query-dependent prompt optimization framework that leverages the expressiveness of textual rewards to directly guide the fine-tuning of a prompt model. Unlike prior work, TRPrompt removes strong dependence on initial expert-provided prompts, enabling prompt learning from scratch through an iterative process guided by textual feedback. Experiments across arithmetic reasoning tasks show that TRPrompt excels on challenging datasets like MATH and GSMHard, where rich textual feedback offers meaningful guidance. These results highlight TRPrompt’s potential for aligning LLMs in settings where numerical rewards are limited or ill-defined, and position textual supervision as a powerful tool for prompt optimization.

\begin{ack}
\end{ack}

\bibliographystyle{plain}
\bibliography{ref}


\appendix

\section{Technical Appendices and Supplementary Material}

\subsection{Synthetic Dataset Creation}
A central component of our pipeline is the construction of synthetic training datasets. At each iteration, we use the current versions of the Prompt Model and Reward Model to generate a new dataset, which is then used to fine-tune the next version of the Prompt Model. This process allows the model to iteratively improve by learning from its own outputs and the corresponding feedback.

Synthetic dataset creation consists of two key steps:
\textbf{(1) Prompt Generation}, where the Prompt Model generates a query-specific prompt for a given question, and
\textbf{(2) Textual Reward Generation}, where the Reward Model produces textual rewards for the generated prompt.

Given that both models rely on LLMs, clear and precise instruction is critical to ensure the correct behavior of each model during this process. To this end, meta-instructions were carefully designed to align each model with its respective objective.

\begin{figure}[htbp]
    \centering
     \includegraphics[width=1\linewidth]{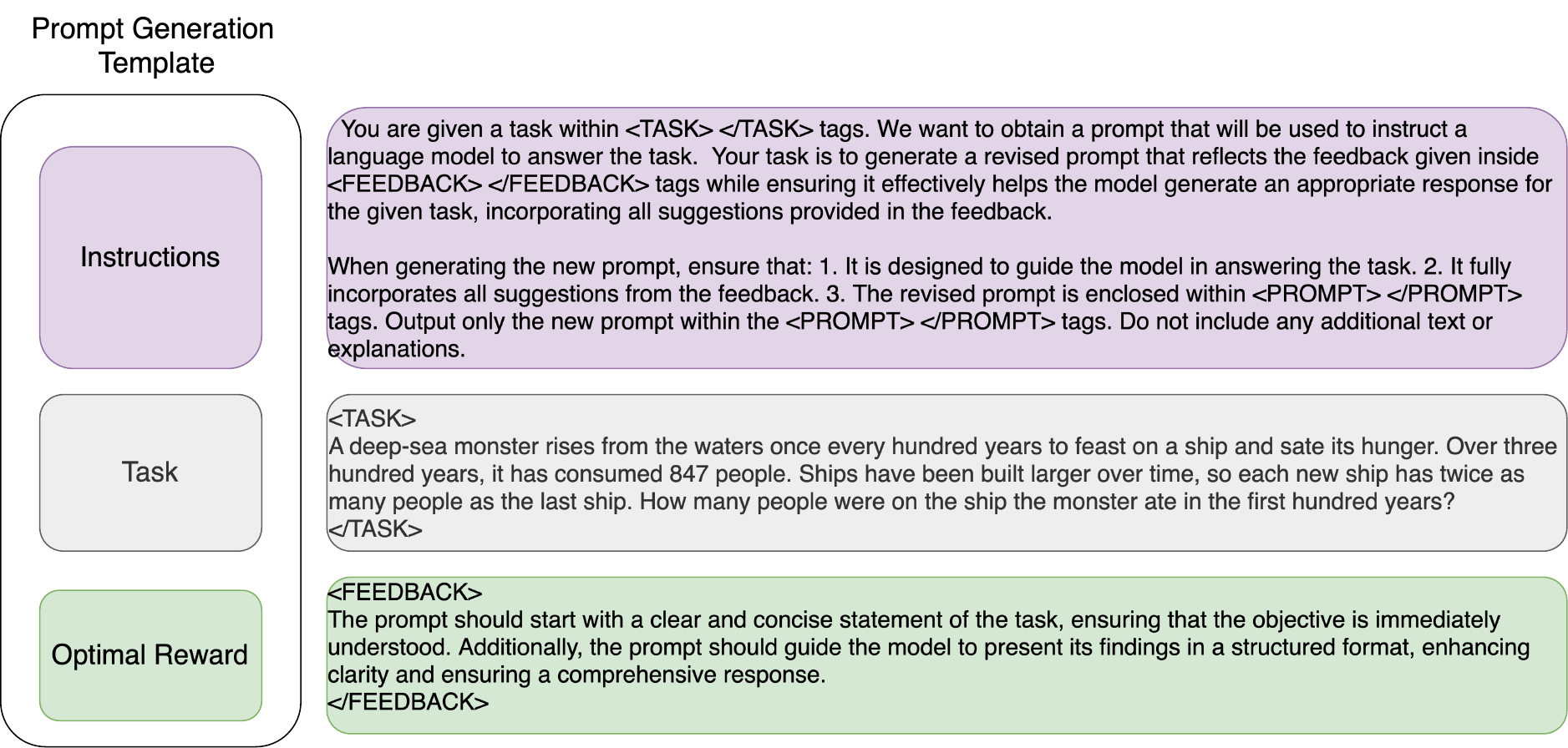}
    \caption{Meta-Instruction template and example used to generate query-dependent prompts conditioned on the Optimal Reward.}
    \label{fig:prompt_template}
\end{figure}

\paragraph{Prompt Generation} To generate prompts conditioned on the optimal textual reward, we instruct the Prompt Model using a structured input format. The meta-instructions specify both the task and the corresponding Optimal Reward, which serves as the guiding feedback. As illustrated in Figure \ref{fig:prompt_template}, the Prompt Model is prompted with a clear definition of its input and output expectations and is explicitly instructed to generate a query-dependent prompt that aligns with the provided textual reward.

\begin{figure}[htbp]
    \centering
    \includegraphics[width=1\linewidth]{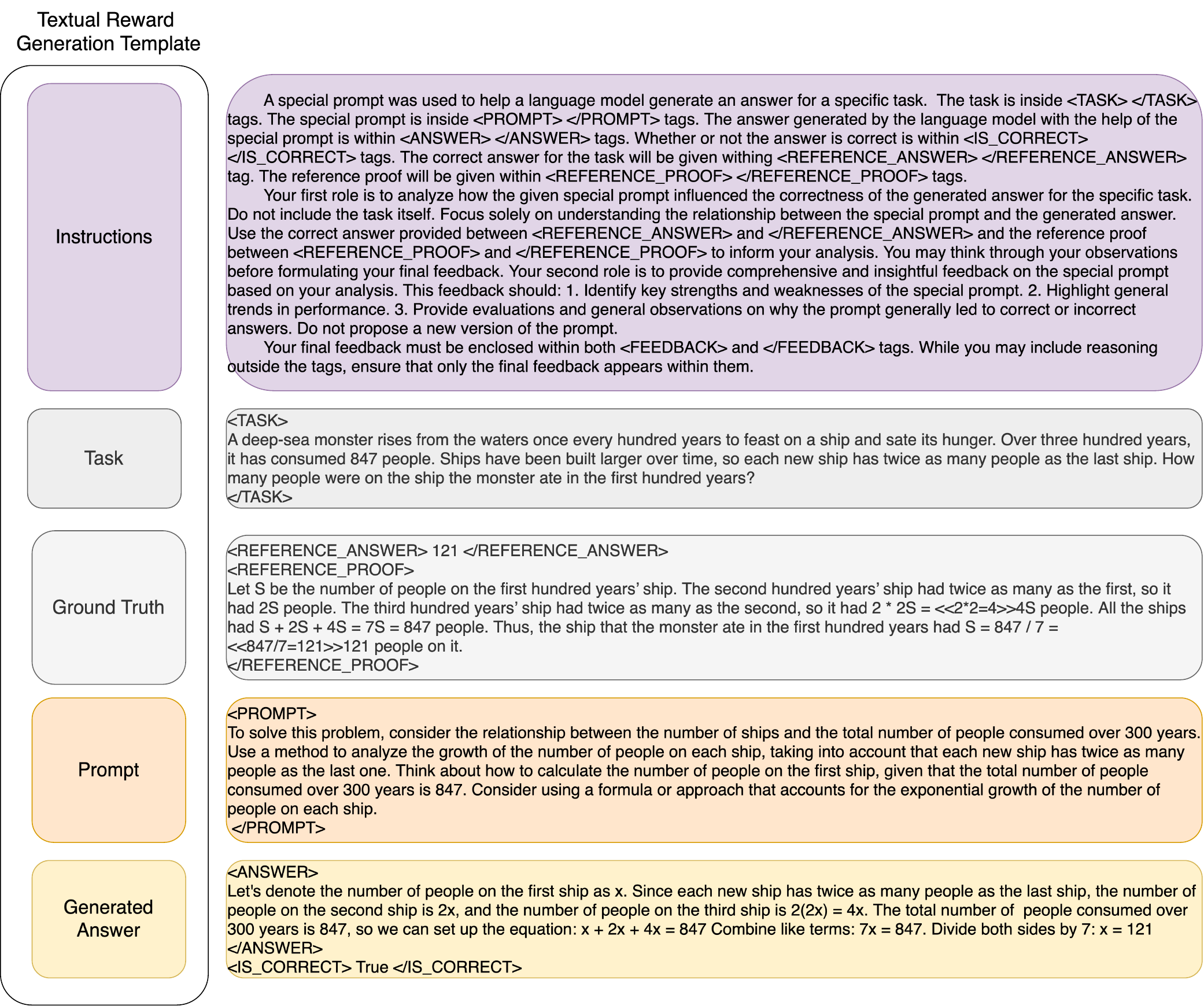}
    \caption{Instruction template and example used to guide the Reward Model in generating a textual reward. The input includes the task, the query-dependent prompt, the ground-truth answer, and the Target Model’s answer to the question using the query-dependent prompt.}
    \label{fig:reward_template}
\end{figure}

\paragraph{Textual Reward Generation} To generate a textual reward for a specific prompt, we provide the Reward Model with the generated prompt, the Target Model’s output for the associated question, and the corresponding ground-truth answer. The Reward Model is guided by a meta-instruction designed to produce a natural language critique that assesses the quality of the prompt based on the accuracy and relevance of the generated answer. This critique is the textual reward for the query-dependent prompt. An example of this meta-instruction template, along with a concrete instantiation, is shown in Figure \ref{fig:reward_template}.

\begin{figure}[htbp]
    \centering
    \includegraphics[width=1\linewidth]{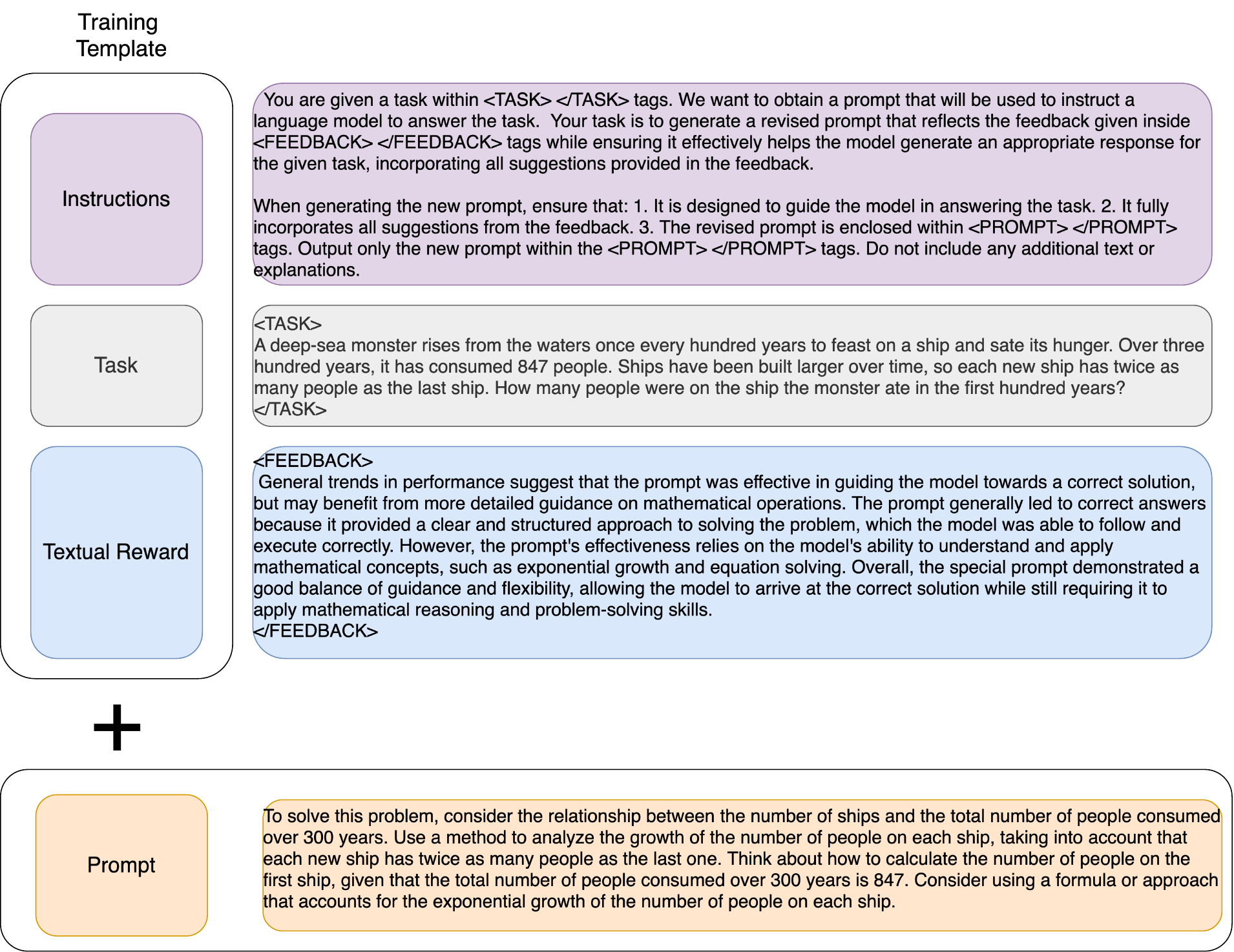}
    \caption{Example of a training pair used in Supervised Fine-Tuning. The Prompt Model is trained to reproduce the query-dependent prompt based on the provided textual reward.}
    \label{fig:training}
\end{figure}

\paragraph{Training Samples} Each synthetic training sample consists of a query-dependent prompt paired with its corresponding textual reward. To incorporate this supervision, we perform Supervised Fine-Tuning (SFT), training the Prompt Model to generate the original prompt conditioned on the textual reward. The model is optimized by minimizing the loss between the generated prompt and the reference prompt that was originally used to compute the textual reward. An example training pair, consisting of a textual reward and its associated prompt, is shown in Figure \ref{fig:training}.

\subsection{Training Details}
\paragraph{Datasets.} For each dataset, we compute test scores on the dedicated test sets, which consist of 1,319 samples for GSM8K, 319 for GSMHard, and 3,369 for MATH. The training sets include 7,473 questions for GSM8K, 1,000 for GSMHard, and 5,136 for MATH. We apply a 90\%–10\% train-validation split on each training set.
\paragraph{Synthetic Dataset.} At each iteration, a synthetic dataset of 800 lengths is created, containing prompts and their corresponding textual rewards. In order to create this dataset, we randomly sample 800 questions from the train set on each iteration. All the final results we report in the paper are on the dedicated test set for each dataset. We use the validation set in order to choose the best checkpoint.
\paragraph{Generations.} When assessing a prompt's performance, we configure the Target LLM's inference temperature to be 0.001 when generating a response to a question concatenated with the query-dependent prompt we wish to evaluate. We also set the temperature to be 0.001 for textual reward generations. We set the default temperature to be 0.9 when generating new query-dependent optimal prompts with the Prompt Model to encourage diverse and creative generation of new prompts for training. We use a batch size of 64 for generations.

\paragraph{Training Details.} We train the Prompt Model for 4 iterations.
\paragraph{SFT.} We train the Prompt Model on the synthetic dataset containing pairs of prompts and their rewards via supervised fine-tuning (SFT). Due to memory and cost restrictions, we choose to use Low-Rank Adaptation (LoRA), a method designed to accelerate fine-tuning while keeping memory consumption low. LoRA decomposes the weights-update-matrix into two lower-dimension matrices that are easier to compute and store in memory. Due to its benefits, LoRa is a suitable choice for optimally fine-tuning the LLM while maintaining computational efficiency and preserving model performance. For LoRA parameters, we utilize \textit{$r=256$} and \textit{$\alpha=256$}. The Prompt Model is trained at each iteration for $2$ epoch with a $2 \times 10^{-5}$ learning rate. We use Adam optimizer, with a linear learning rate decay.

\paragraph{Textgrad} For Textgrad, we use GPT-4o-mini to make the changes to the prompt. Textgrad is run for 10 iterations and uses the validation set in order to compare the quality of the Optimal Reward.

\paragraph{Hardware Requirements.} The training was performed on  NVIDIA A100 GPUs with 80 GB of RAM. The training is done using GPU memory. Training time is between 48h - 72h depending on the dataset. The bottleneck of the training was the Textgrad step, which is not parallelizable. Identifying the Optimal Reward at each iteration accounted for around 70\% of the total training time, primarily due to the non-parallelized evaluation step in Textgrad, which significantly slowed down the training process.


\end{document}